\documentclass[letterpaper, 10 pt, conference]{ieeeconf}  

\IEEEoverridecommandlockouts                              

\usepackage{graphicx} 
\usepackage{mathptmx} 
\usepackage{times} 
\usepackage{amsmath} 
\usepackage{amssymb}  
\usepackage{textcomp}
\usepackage{booktabs}
\usepackage{multirow}
\usepackage{balance}
\usepackage{soul,color}
\usepackage{comment}
\usepackage{array}
\usepackage{cite}
\usepackage[ruled,vlined]{algorithm2e}

\def\Ali#1{{\color{red}{\bf [Ali:} {\it{#1}}{\bf ]}}}

\title{\LARGE \bf
Video-Based Inpatient Fall Risk Assessment: A Case Study
}

\author{
Ziqing Wang$^{1,2}$,
Mohammad Ali Armin$^{1}$, 
Simon Denman$^{3}$,
Lars Petersson$^{1}$, 
David Ahmedt-Aristizabal$^{1,3}$
\thanks{$^{1}$ CSIRO, DATA61, Canberra, Australia. 
{Corresponding author: \tt\footnotesize david.ahmedtaristizabal@data61.csiro.au}
}%
\thanks{$^{2}$ Australian National University, Canberra, Australia 
}
\thanks{$^{3}$ SAIVT Research lab, Queensland University of Technology, Brisbane, Australia.%
}%
}

\begin{document}
\def\Ali#1{{\color{blue}{\bf [Ali:} {\it{#1}}{\bf ]}}}

\maketitle
\thispagestyle{empty}
\pagestyle{empty}

\begin{abstract}
Inpatient falls are a serious safety issue in hospitals and healthcare facilities. Recent advances in video analytics for patient monitoring provide a non-intrusive avenue to reduce this risk through continuous activity monitoring. However, in-bed fall risk assessment systems have received less attention in the literature. The majority of prior studies have focused on fall event detection, and do not consider the circumstances that may indicate an imminent inpatient fall.
Here, we propose a video-based system that can monitor the risk of a patient falling, and alert staff of unsafe behaviour to help prevent falls before they occur.
We propose an approach that leverages recent advances in human localisation and skeleton pose estimation to extract spatial features from video frames recorded in a simulated environment.
We demonstrate that body positions can be effectively recognised and provide useful evidence for fall risk assessment.
This work highlights the benefits of video-based models for analysing behaviours of interest, and demonstrates how such a system could enable sufficient lead time for healthcare professionals to respond and address patient needs, which is necessary for the development of fall intervention programs.
\end{abstract}

\section{INTRODUCTION}

Falls in the ward, in particular those from the bed, are a persistent problem and are commonly associated with injuries such as soreness and bone fractures and often result in a prolonged hospital stay. In mental health hospitals and some psychogeriatric units, these events are of particular concern due to patient cognitive impairment, dizziness or vertigo~\cite{oliver2010preventing,vassallo2000epidemiological}. Such incidents are one of the main concerns for all staff involved in the care of patients, and can lead to anxiety or guilt, and potentially litigation.
In most hospitals, medical staff follow well-defined protocols to prevent falls, however research into systems capable of generating immediate alerts to enable medical assists to prevent falls has received limited attention from researchers.

Considering the importance of patient behaviour monitoring, several in-clinic patient monitoring systems using computer vision and deep learning have been introduced to provide an objective assessment of a patient's behaviour. 
These vision-based systems have attracted great attention due to their non-invasive nature and have shown promising results in analysing patient-specific pose~\cite{chen2018patient} (for example, sleeping pose~\cite{liu2019seeing,liu2019bed}), epileptic patients~\cite{ahmedt2019understanding}, breathing disorders~\cite{martinez2019vision} and infant motions~\cite{hesse2019learning}.
%
Furthermore, camera-based fall detection~\cite{lu2018deep,chen2020vision,asif2020privacy}
and fall prediction systems~\cite{hua2019falls,masalha2020predicting}, 
which detect when a fall occurs rather than seeking to predict a fall before it happens, have also received considerable attention recently
and have achieved effective results using existing simulated datasets~\cite{auvinet2010multiple,charfi2013optimized,kwolek2014human}, or synthetic libraries~\cite{asif2020sshfd,asif2020privacy}.
However, while human pose estimation has become the de-facto standard for inpatient analysis, its application to the prevention of falls from the bed remains limited. 

To prevent falls, some systems detect the position in which the patient is lying with respect to the edge of the bed, or detect the patient's bed-exit behaviour. In these scenarios, the system monitors a key human pose or human motion to predict the risk of falling.
These studies have used commercial pressure mat systems to detect the human off-bed position~\cite{viriyavit2020bed}; however, pressure pads have been shown to generate a high volume of false alarms leading to alarm fatigue~\cite{shorr2012effects}. 
On the other hand, camera-based systems seek to detect a sitting posture~\cite{inoue2019bed} or the bed exit action from a sequence of human images~\cite{inoue2020bed}.
Although, many hospital patients fall as they get out of bed, there are risk factors for falling such as uncontrolled motions caused by agitations, restless sleep and abnormal dreams that lead to a patient trying to climb out of the bed for protection, or "jumping" from the bed. 
Patients attempting to perform these activities unassisted account for a large proportion of inpatient falls, and they are the focus of this paper.

\begin{figure}[!t]
\centering
\includegraphics[width=0.92\linewidth]{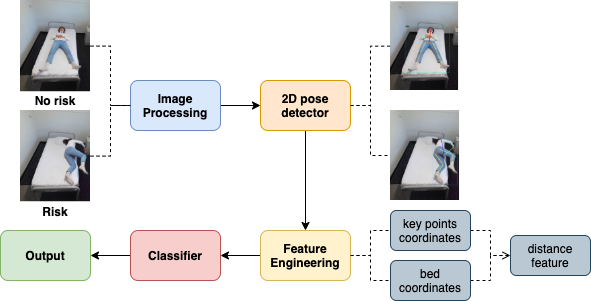}
\vspace{-4pt}
\caption{
Overview of the proposed fall risk monitoring system. Given a single RGB image collected with a custom camera system placed in the ceiling, the system generates 2D human pose predictions. Next, the relative position of the human and the bed is computed to predict the risk of falling. }
\label{fig:Fig1}
\vspace{-13pt}
\end{figure}

In this paper, we explore the feasibility of adapting pose-based frameworks to identify patients' behaviour and assess the risk of falling from the bed.

Our main contributions are summarized as follows:
\begin{enumerate}
\vspace{-2pt}
\item We introduce a flexible vision-based fall risk detection system capable of detecting actions in a novel simulated environment that may indicate an imminent inpatient fall.
\item We propose a robust but simple non-obtrusive monitoring system to capture relative body position information to assess the risk of falling from a bed. 
\end{enumerate}
\vspace{-3pt}


\begin{figure}[!t]
\centering
\includegraphics[width=0.6\linewidth]{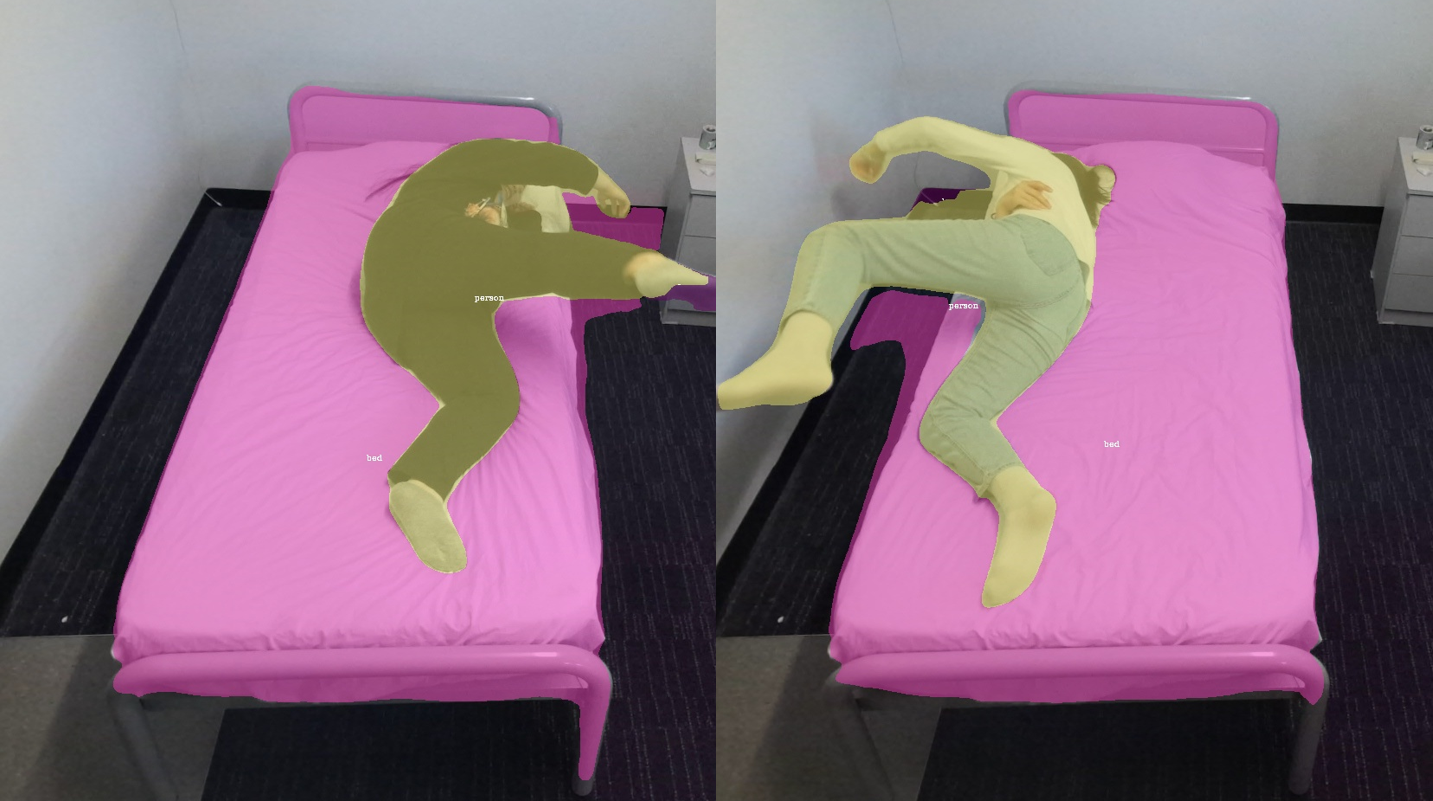}
\vspace{-4pt}
\caption{
Human and bed localisation. A dynamic and fast instance segmentation approach is used to localise the region of interest.
}
\label{fig:Fig3}
\vspace{-4pt}
\end{figure}

\begin{figure}[!t]
\centering
\includegraphics[width=1\linewidth]{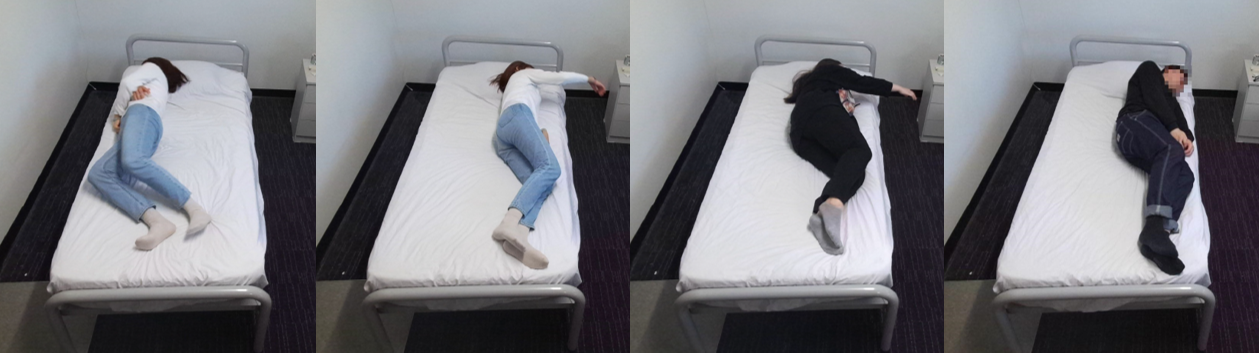}
\includegraphics[width=1\linewidth]{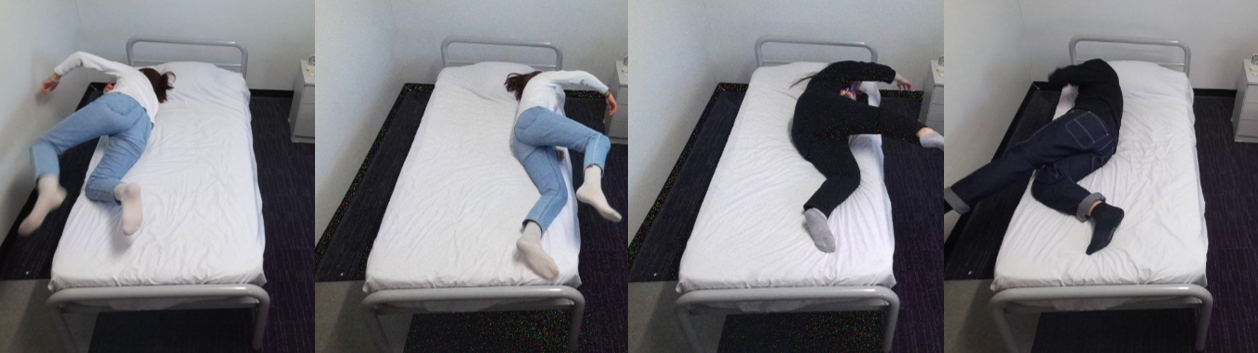}
\vspace{-16pt}
\caption{
Selected samples of human actions collected in the simulated dataset. Top: Not at risk of falling. Bottom: At risk of falling.
}
\label{fig:Fig2}
\vspace{-6pt}
\end{figure}

\vspace{-4pt}
\section{MATERIALS AND METHODS}

In this paper, we propose a non-invasive landmark-based approach to capture in-bed human pose, and predict whether the human in an image will fall from the bed they are currently occupying.

\subsection{Fall prevention simulated dataset and pre-processing}
Existing vision-based fall detection datasets do not cover inpatient fall events, or examples of patients at risk of falling. Considering this limitation, we designed and collected a simulated dataset with the actions of interest represented by two classes, \textit{not at risk} and \textit{at risk} of falling. 
To generate this data two main stages were covered: i) recruitment, experimental setup, and data collection, and ii) data annotation and pre-processing. 

We collect data from participants in a simulated hospital environment. Volunteers lie down on the bed and simulate in-bed patient actions such as trying to climb out of the bed for protection, turning around, exiting the bed, and falling. All images and videos are collected with a custom recording device equipped with a Microsoft Azure Kinect camera, and are saved for a further prepossessing phase.

To estimate relative positions of the human and bed, we first define the region of interest as the location of the bed that contains the participant. This process helps to deal with different camera-bed viewing angles, and changes in the inclination angle of the bed which may impact the risk assessment. 
We perform object boundary detection using SOLO~\cite{wang2020solov2}, which is a dynamic and fast state-of-the-art instance segmentation method. In our network, we use pre-trained weights, trained on the COCO dataset~\cite{lin2014microsoft}, to detect the human and the bed.
Then, we crop and resize all images to a resolution of $1080\times828$ pixels as input to the system. Fig.~\ref{fig:Fig3} depicts selected examples of the participant and bed detection. 
Finally, each image is separated into two classes to conduct the proposed analysis: \textit{at risk} and \textit{not at risk} of falling from the bed, as illustrated in Fig.~\ref{fig:Fig2}. We address the data imbalance in our dataset by adopting data augmentation techniques including oversampling and adding Gaussian noise to images to  obtain the same number of samples for both classes. We argue that this pre-processing step does not impact generalization.

\begin{figure}[!t]
\centering
\includegraphics[width=0.8\linewidth]{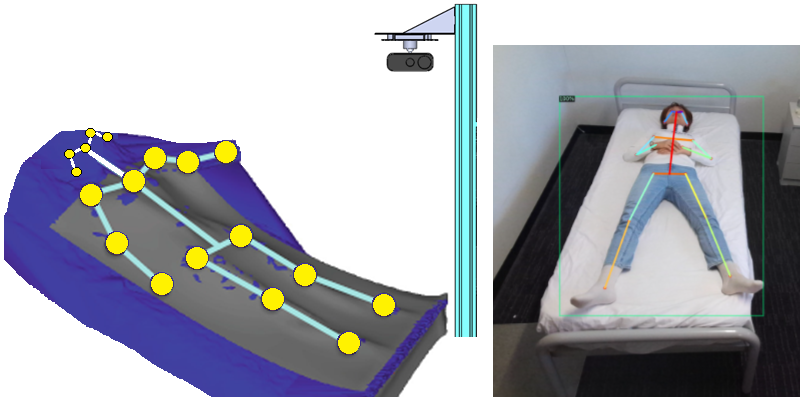}
\vspace{-10pt}
\caption{
Representation of the human key points detected in a selected image.
}
\label{fig:Fig4}
\vspace{-13pt}
\end{figure}

\subsection{Fall risk assessment system}
Fig.~\ref{fig:Fig1} shows the overall architecture of our framework which has three main modules: \textbf{i}) 2D key-point estimation, which takes an RGB image and produces body joint locations in 2D space; \textbf{ii}) A human-bed relative position estimation and feature engineering; and \textbf{iii}) a fall risk classifier which combines human pose and relative position features to accurately discriminate between \textit{at risk} and \textit{not at risk} cases. In the following, we describe in detail the individual components of our framework.

\subsubsection{Human localisation and pose identification}
Quantifying a person's posture and limb articulation is useful for understanding patient behaviour. Human pose estimation from static images has shown strong performance in detecting positions of interest for the analysis of seizure disorders~\cite{ahmedt2019understanding} and bed-exit posture~\cite{inoue2019bed}.

We aim to employ a robust 2D pose prediction technique to extract consistent poses from a hospital environment, where challenges such as self-occlusion and similarities between the  background and foreground are present. 
We adopt the Mask-RCNN architecture~\cite{he2017mask} to predict 2D locations of body joints and their corresponding confidence scores, which is a lightweight, yet highly effective approach implemented in Detectron2~\cite{wu2019detectron2}. Here, a keypoint location is modeled as a one-hot mask where Mask-RCNN predicts $k$ masks, one for each of $K$ keypoints (17 key points coordinates in this study). We fine tune a pre-trained Mask-RCNN model trained on the COCO dataset~\cite{lin2014microsoft}, enabling the model to detect keypoints clearly in our dataset. 
When participants perform activities such as turning around, their body parts may overlap and some key points cannot be detected clearly. Based on an analysis of the pose estimation results, we define the most stable coordinates for a later feature engineering step.
Fig.~\ref{fig:Fig4} illustrates the keypoint layout and a detected human pose.

\begin{figure}[!t]
\centering
\includegraphics[width=0.46\linewidth]{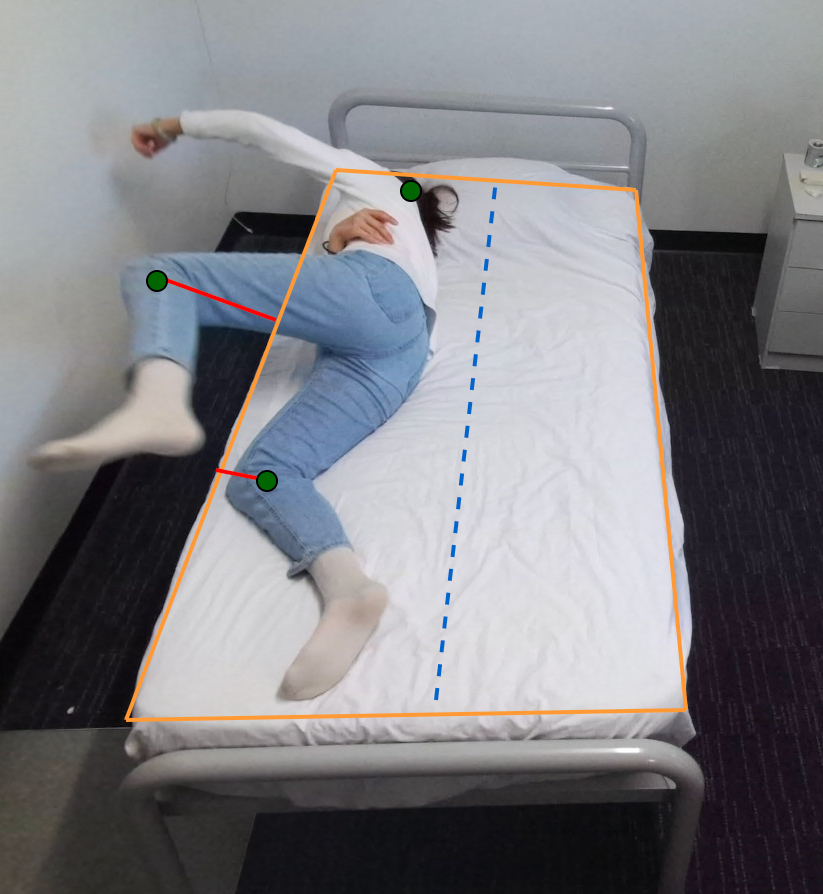}
\vspace{-5pt}
\caption{
Representation of the distance feature estimation. Head and knees detected by the pose estimation algorithm are marked as green dots, and all lines represent the contour of the detected bed.
}
\label{fig:Fig5}
\vspace{-6pt}
\end{figure}

\begin{table}[t!]
\caption{Multi-Fold Cross-Validation Performance (10-Time Average)}
\vspace{-10pt}
\centering
\resizebox{0.47\textwidth}{!}{
\begin{tabular}{
lc|c
}
\toprule
  & \multicolumn{2}{c}{\textbf{Test Accuracy (\%)}}   \\
\textbf{Feature representation} & \textbf{Light GBM} &  \textbf{SVM}   \\
\midrule
$Dis<Knee,Bed>$                               & 94.44  &  91.11 \\
$Dis<Knee,Bed> + Dis<Head,Bed>$               & 96.11  &  92.89 \\
$17 Key points + Dis<Knee,Bed>$               & 96.67  &  93.89 \\
$17 Key points + Dis<Knee,Bed> + Dis<Head,Bed>$  & \textbf{97.22}  &  94.20 \\
\bottomrule
\vspace{-16pt}
\end{tabular}}
\label{tab:results} 
\end{table}

\begin{figure}[!t]
\centering
\includegraphics[width=1\linewidth]{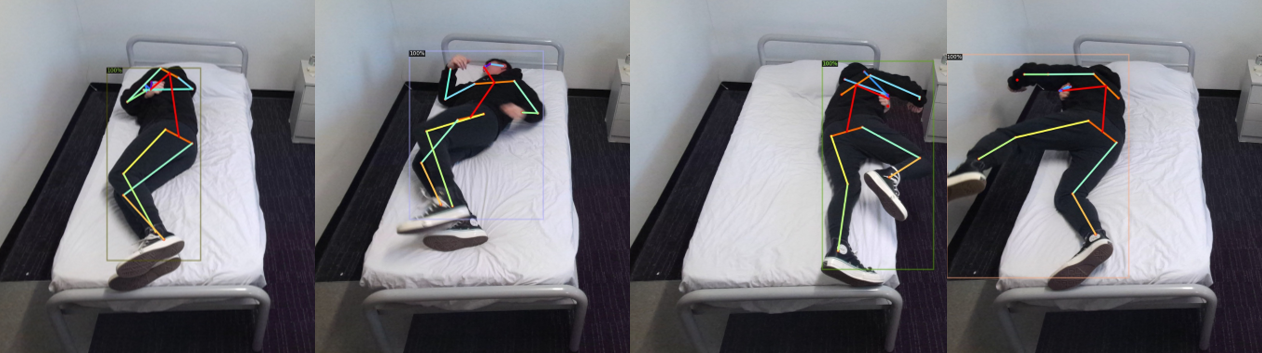}
\vspace{-15pt}
\caption{
Qualitative results of human pose estimation on patients \textit{not at risk} and \textit{at risk} of falling (trying to climb out of the bed for protection).
}
\label{fig:Fig6}
\vspace{-13pt}
\end{figure}

\subsubsection{Relative position determination and feature engineering}

Human body postures during events such as the patient climbing out of the bed are complex and varied, so it is essential to define a criteria to classify whether a patient is at risk of falling from the bed or not. One criteria is to track and estimate how much of the knee is outside of the bed. 

To calculate the distance between the knee and the bed, we need to determine which side of the bed a patient is most likely to exit from.
As shown in Fig.~\ref{fig:Fig5}, the left line, middle line and right line of the bed define the bed position, and the head and knees are marked as green dots. 
When the head and the two knees are all on the left of the middle line, the human body is defined as being on the left side of the bed. When the head and two knees are all on the right of middle line, the human body is defined as being on the right side of the bed. If the human body is on the left side of the bed, we will calculate the distance between the two knees and the left line of the bed. If the human body is on the right side of the bed, we will calculate the distance between the two knees and the right line of the bed. 
Defining where the human body is located can also help to remove unnecessary features and decrease feature dimensionality. 

The distance feature is the distance between the knees and the bed boundary. Based on the human body's location on the bed, we can determine which boundary should be used to calculate the distance between the knee and the bed. If the knee is outside of the bed, the distance value is negative and if the knee is inside the bed, the distance value is positive. 

\subsubsection{Fall risk classification}

Each output feature related to the human pose coordinates and relative position between the body and the bed (i.e. distance features) is fed to a classifier to learn probabilistic distributions with respect to the target class.
We adopt the LightGBM classifier~\cite{ke2017lightgbm} as a fast, distributed, high performance implementation of gradient boosted trees for supervised classification with robustness to overfitting. 
The following setting is used in the experiment: boosting type (gradient boosting decision tree), boosting learning rate (0.1), number of boosted trees to fit (100), maximum tree leaves (31), maximum tree depth (no limit).
We also use a traditional support vector machine (SVM) classifier with automatic Bayesian optimization: kernel (sigmoid), shrinking (true), cache size (200).

\vspace{-2pt}
\section{EVALUATION}

\subsection{Experimental setup}

To evaluate and compare the most discriminative features from the landmark-based analysis, we adopt four feature sets: \textbf{i)} the distance between knees and the bed; \textbf{ii)} set (i) plus the distance between the head and the bed; \textbf{iii)} set (i) plus 17 body keypoint coordinates; and \textbf{iv)} set (iii) plus the distance between the head and the bed. These feature sets are listed in Table~\ref{tab:results}.

As an ablation study, we investigate a region-based approach by training a ResNet50~\cite{he2016deep} architecture to extract a spatial representation directly from images, and perform classification using a fully connected layer with a sigmoid activation function. This model was implemented in Keras~\cite{chollet2017keras} and trained by optimizing the categorical cross-entropy loss with the Adam optimizer~\cite{kingma2014adam}.

Both models (landmark-based and region-based), are assessed through a 10-fold cross validation to ensure that the training and test data is disjoint. For each fold, the data sample of each class is randomly split into 90\% for training and 10\% for testing.

\vspace{-4pt}
\subsection{Experimental results and discussion}
While general pose estimation frameworks are effective when subjects are located in uncluttered settings, they can be unreliable when applied to noisy environments such as patient monitoring rooms. This can cause even more confusion when multiple frames are fed into the framework, thus in our study we consider a static scene.
From the pose estimation results on the dataset, we confirm that knee points are the most stable keypoints, so we choose to use the knee as a reference to classify \textit{at risk} events and use the head location to identify the human position on the bed. Selected samples showing estimated joint locations are presented in Fig.~\ref{fig:Fig6}. 

The region-based approach achieved an average accuracy of 65.8\% on the test set, whereas the landmark-based approach using LightGBM achieved 97.22\% accuracy. The cross-validation performance for each set of features and the proposed classifiers are shown in Table~\ref{tab:results}. 
Our landmark-based results indicate that larger feature sets can improve system performance. The best accuracy is obtained from the fourth feature set using 17 keypoint coordinates, the distance between the knee and the bed and the distance between the head and the bed. This performance gain is likely a result of the distance features indicating whether key body parts overlap the boundary of the bed, which indicates a high probability of falling from the bed.
This relative position information is hard to capture without feature engineering and is the main reason that the region-based approach shows low performance.

In most hospitals, there are programs and policies for fall prevention, but there is limited research into systems capable of generating immediate alerts for medical assistance to prevent falls. A video-based alarm system for fall prevention similar to our proposed framework needs the ability to detect highly accurate relative human position in the bed. Such a system is able to issue an alert as early as possible once it detects a position where there is a high risk of falling. Further, it is envisioned that our approach is cost effective and low maintenance. 

The majority of inpatient fall studies focus solely on fall risk factors but may not identify potential causal factors for falls (e.g. what triggered the fall) which is necessary for fall intervention programs. 
An interesting direction for future research is the creation of libraries of behaviours to identify these factors and patients at high risk of falling in the early stages of monitoring.

\section{CONCLUSIONS}
In this paper, we introduce a vision based monitoring system that incorporates state-of-the-art computer vision techniques to assess the risk of falling from a bed. Considering the lack of datasets to assess fall risk and fall prevention, we introduce a simulated dataset that includes in-bed human actions such as trying to climb out of the bed for protection, turning around, and bed-exit events. Our results in this particular case study show a promising technology that can have a positive impact on monitoring inpatients at risk of falling. Our proposed system has a high accuracy, resulting in lower false alarm rates for medical staff and thus a reduction in the likelihood of alarm fatigue. 

\paragraph*{Ethics statement}
The experimental procedures involving human subjects described in this paper were approved by the CSIRO Health and Medical Human Research Ethics Committee (CHMHREC). 

\balance


\bibliographystyle{IEEEtran}
\bibliography{refs}

\vspace{0.5cm}

\end{document}